\documentclass[letterpaper, 10 pt, conference]{ieeeconf}  

\usepackage{breakurl}
\usepackage{graphicx}
\usepackage{amsfonts}
\usepackage{booktabs}
\usepackage{multirow}
\usepackage{hyperref}
\let\labelindent\relax
\usepackage{enumitem}
\usepackage{amsmath}
\usepackage{xcolor}
\usepackage{bbding}
\IEEEoverridecommandlockouts                              

\title{\LARGE \bf
GS-LTS: 3D Gaussian Splatting-Based Adaptive Modeling\\ for Long-Term Service Robots
}

\author{Bin Fu$^{1}$ and Jialin Li$^{1}$ and Bin Zhang$^{1}$ and Ruiping Wang$^{1,}$\textsuperscript{\Envelope} and Xilin Chen$^{1}$
\thanks{*This work is partially supported by National Key R\&D Program of China No. 2021ZD0111901, and Natural Science Foundation of China under contracts Nos. 62495082, U21B2025.}
\thanks{$^{1}$The authors are with the Key Laboratory of AI Safety of CAS, Institute of Computing Technology, Chinese Academy of Sciences (CAS), Beijing, 100190, China, and also with the University of Chinese Academy of Sciences, Beijing, 100049, China. {\tt\small \{bin.fu, jialin.li, bin.zhang\}@vipl.ict.ac.cn}, \tt\small \{wangruiping, xlchen\}@ict.ac.cn}
\thanks{\textsuperscript{\Envelope}Corresponding author.}
}

\begin{document}

\maketitle
\thispagestyle{empty}
\pagestyle{empty}

\begin{abstract}

3D Gaussian Splatting (3DGS) has garnered significant attention in robotics for its explicit, high fidelity dense scene representation, demonstrating strong potential for robotic applications.
However, 3DGS-based methods in robotics primarily focus on static scenes, with limited attention to the dynamic scene changes essential for long-term service robots.
These robots demand sustained task execution and efficient scene updates—challenges current approaches fail to meet.
To address these limitations, we propose GS-LTS (Gaussian Splatting for Long-Term Service), a 3DGS-based system enabling indoor robots to manage diverse tasks in dynamic environments over time.
GS-LTS detects scene changes (e.g., object addition or removal) via single-image change detection, employs a rule-based policy to autonomously collect multi-view observations, and efficiently updates the scene representation through Gaussian editing.
Additionally, we propose a simulation-based benchmark that automatically generates scene change data as compact configuration scripts, providing a standardized, user-friendly evaluation benchmark.
Experimental results demonstrate GS-LTS’s advantages in reconstruction, navigation, and superior scene updates—faster and higher quality than the image training baseline—advancing 3DGS for long-term robotic operations.
Code and benchmark are available at: \url{https://vipl-vsu.github.io/3DGS-LTS}

\end{abstract}


\section{Introduction}

\begin{figure}[t]
\centering
\includegraphics[width=1\linewidth,height=0.60\linewidth]{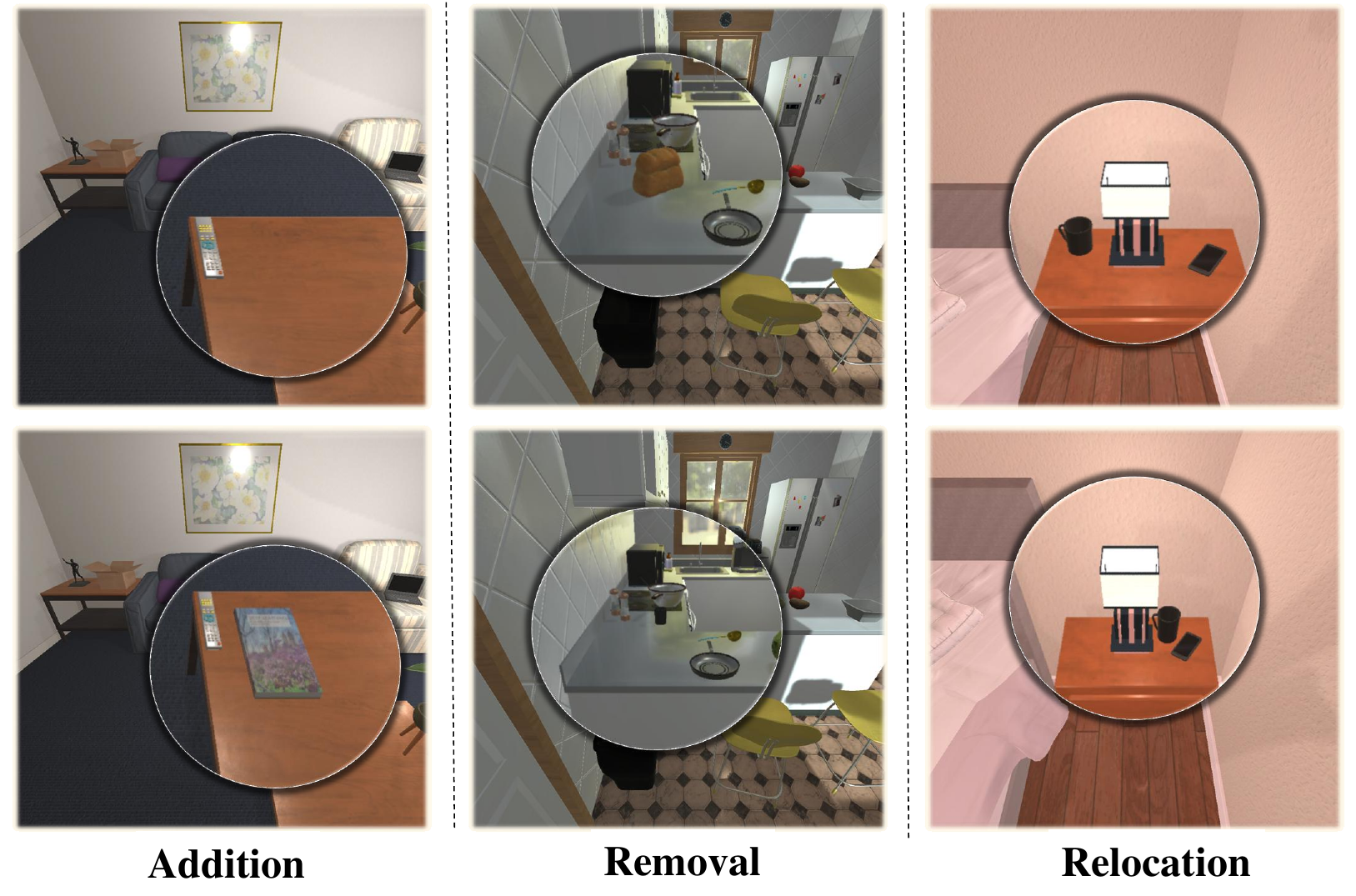}
\vspace{-1.8em}
\caption{Three common types of scene changes in indoor scenes.}
\label{fig:head}
\vspace{-1.7em}
\end{figure}

3D Gaussian Splatting (3DGS) \cite{kerbl20233d} is an explicit radiance field representation based on 3D Gaussians. It has been widely applied in fields such as dense visual SLAM \cite{keetha2024splatam} and 3D reconstruction \cite{wu2024hgs}, benefiting from its explicit geometric structure and real-time high-quality rendering.
By further embedding low-dimensional vision-semantic features into each 3D Gaussian \cite{langsplat}, a comprehensive scene representation integrating geometry, vision, and semantics can be achieved, which shows great potential in robotics applications, such as navigation and instruction following.
However, current 3DGS attempts in these fields primarily focus on static scenes \cite{zhu20243d}, which fail to align with the dynamic nature of real-world environments involving object changes, as illustrated in Fig. \ref{fig:head}, making these approaches inadequate for long-term service robots working in dynamic settings.
A more realistic scenario involves a robot utilizing a prebuilt 3DGS representation to perform tasks in an environment where objects may be added, removed, or relocated over time.
In such environments, the robot must continuously observe the scene, autonomously detect changes, and update its scene representation to maintain accuracy.

A straightforward approach to handling scene changes would be to periodically recollect images and retrain or fine-tune the 3DGS representation whenever the environment is modified. However, this method is computationally expensive, requiring frequent reprocessing of large-scale data, and lacks efficiency for real-time or long-term deployment.
To address this task, we propose \textbf{GS-LTS}, a 3D\textbf{GS}-based system designed for \textbf{L}ong-\textbf{T}erm \textbf{S}ervice robots in indoor environments. The GS-LTS framework integrates four key modules: (1) \textbf{Gaussian Mapping Engine}, which constructs a semantic-aware 3DGS representation, integrating geometry, visual appearance, and semantics; (2) \textbf{Multi-Task Executor}, which helps robots perform downstream tasks like object navigation using the informative 3DGS representation; (3) \textbf{Change Detection Unit}, a long-running module that detects scene changes at a specified frequency by comparing the robot’s current RGB observations with historical 3DGS-rendered images, locating altered regions and analyzing change types and positions; and (4) \textbf{Active Scene Updater}, which is guided by a rule-based policy, directs the robot to collect multi-view images around detected areas, and applies pre-editing and fine-tuning to dynamically update the 3DGS representation based on the detect change type and new observations.
Together, these components enable the robot to adapt to evolving surroundings while maintaining robust performance over extended periods.


Evaluating this system places significant demands on both data availability and environmental support.
Real-world settings struggle to support both robotic task execution and extensive variations, hindering large-scale dataset creation and standardized evaluation.
To address this, we propose a simulation-based benchmark that supports task execution and policy learning via 3DGS representations while enabling systematic generation of large-scale scene change data through object interactions.
This benchmark not only facilitates large-scale evaluation but also serves as a bridge for sim-to-real transfer, allowing models trained in simulation to achieve enhanced performance in real-world environments.
Our approach features two innovations: (1) automated generation of customizable scene change data, combining objects (e.g., cups), containers (e.g., tables), and positions to produce diverse scene change tasks; and (2) storing scene change setups and environment metadata in configuration scripts, which ensures efficient storage, easy configuration, and accurate reproduction of scenes.
This scalable, reproducible benchmark reduces data acquisition costs and provides standardized evaluation, advancing research on 3DGS adaptability in dynamic environments.

We conduct extensive validation of the GS-LTS system through a series of experiments.
First, we evaluate scene representation quality via image rendering for visual fidelity and 3D localization for semantic accuracy.
Additionally, object navigation results on an existing benchmark \cite{wortsman2019learning} highlight the potential of 3DGS for embodied tasks.
Finally, on our custom Scene Change Adaptation Benchmark, we compare our Gaussian editing-based method with the baseline of direct image fine-tuning.
Our approach significantly reduces scene update time while enhancing update quality.
These comprehensive experiments fully demonstrate the efficiency and robustness of the GS-LTS system in scene reconstruction, embodied applications, and scene adaptability.


In summary, this work introduces GS-LTS, delivering three key contributions:
\begin{itemize}
\item A 3DGS-based system enabling indoor robots to handle diverse tasks in dynamic environments over time.
\item An automatic framework for object-level change detection and adaptive scene update via Gaussian editing.
\item A scalable method for constructing a simulation benchmark for object-level scene change detection.
\end{itemize}
Together, these advancements enhance 3DGS applications for long-term robotic operations in dynamic environments.


\section{Related Work}

\subsection{3D Scene Representation}

Building accurate scene representations is crucial for robotics, with various methods, such as semantic maps~\cite{chaplot2020object}, SLAM~\cite{chen2019suma++}, and NeRF~\cite{mildenhall2021nerf} being widely used. Compared to these representations, 3D Gaussian Splatting (3DGS) provides an explicit, high-fidelity, and real-time renderable dense representation. Its ability to simultaneously encode geometric, visual, and semantic information has driven its adoption in tasks such as 3D reconstruction~\cite{wu2024hgs}, 3DGS-based SLAM~\cite{keetha2024splatam}, and navigation~\cite{jin2024gs}. However, most existing applications are restricted to static environments~\cite{zhu20243d}, where maps quickly become outdated in the face of scene changes.

A key advantage of 3DGS is its inherently editable nature, enabling dynamic updates through direct modifications of Gaussians~\cite{chen2024gaussianeditor}. This adaptability makes 3DGS suited for modeling dynamic environments.
Leveraging these properties, this work explores the integration of 3DGS with long-term service robot systems operating in dynamic settings.


\subsection{Long-Term Robot Autonomy and Change Detection}


Long-Term Autonomy (LTA) is a critical research area in robotics, aimed at enabling robots to operate reliably in complex environments over extended periods~\cite{kunze2018artificial}.
This capability is essential across various domains, including underwater exploration~\cite{jones2012slocum}, and service robotics~\cite{hawes2017strands}.
A major challenge in LTA is adapting to scene changes.
Our work focuses on medium-term changes~\cite{kunze2018artificial} in indoor service environments, where robots must effectively model and update representations of daily object variations.
While many LTA robotic systems have been deployed in service scenarios~\cite{hawes2017strands,hanheide2017and}, our work introduces a novel approach leveraging 3DGS for scene representation to enable efficient adaptation to dynamic environments.

Scene change detection is a key research area in computer vision, aiming to identify scene changes such as object appearance, disappearance, or modifications.
It is broadly classified into 2D and 3D approaches based on data type.
2D change detection employs pairs of before-and-after RGB images~\cite{alcantarilla2018street}, leveraging models from CNNs to foundation models for feature extraction and change identification.
Conversely, 3D change detection incorporates spatial information, relying on multi-view RGB images~\cite{palazzolo2018fast} or point clouds~\cite{wald2019rio}.
Recent advances in 3DGS-based novel-view synthesis \cite{lu20253dgs} have demonstrated strong potential, 
whereas our GS-LTS system adopts a distinct approach, leveraging a single egocentric RGB image for change detection to reduce data and computational demands.


\begin{figure*}[ht]
\centering
\includegraphics[width=1\linewidth]{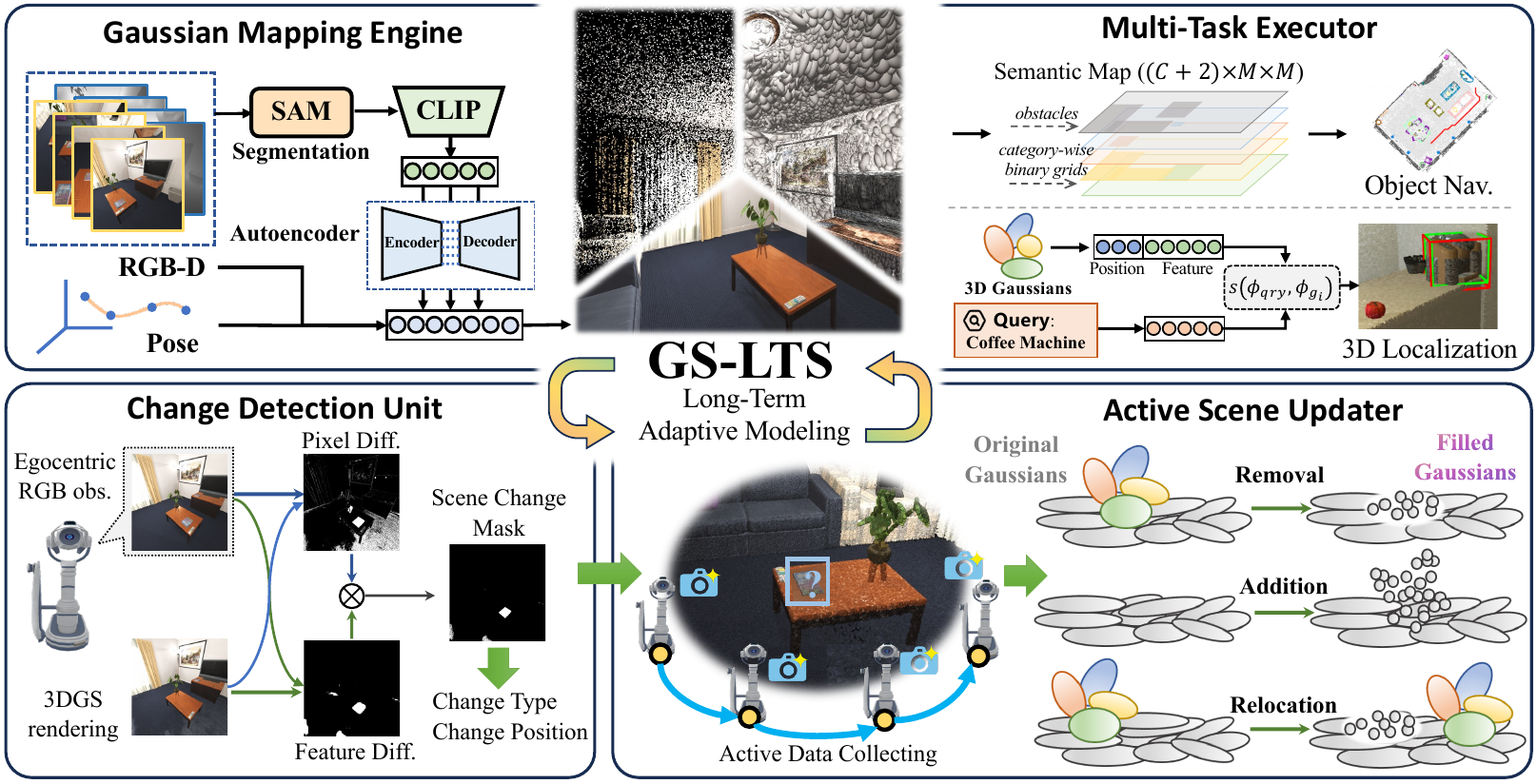}
\vspace{-1.8em}
\caption{\textbf{System Overview}. GS-LTS is a modular system designed for long-term service robots, which can adapt to object changes in the dynamic environments and update the 3DGS representation through periodic, automated operation of the Change Detection Unit and the Active Scene Updater.}
\label{fig:main}
\vspace{-1.6em}
\end{figure*}

\section{System Overview}
\label{System-Overview}
In this section, we first introduce the task formulation for long-term service robot system working in dynamic environments based on 3D Gaussian Splatting (3DGS) scene representation. Subsequently, we present the overall framework of our proposed system designed to address this task.

\subsection{Task Formulation}
The core objectives of the task are twofold: (1) in a dynamic environment, construct a 3DGS representation of the scene and utilize this representation to control the robot in accomplishing downstream embodied tasks, such as object navigation; (2) enable the robot to detect object changes in the scene and autonomously collect data to update the 3DGS representation.

We focus on dynamic indoor settings where primary structures (e.g., room layouts, large furniture like cabinets and refrigerators) stay static, while certain objects (e.g., cups, laptops) exhibit periodic changes.
We consider three types of scene changes: \textbf{relocation}, \textbf{addition}, or \textbf{removal} of objects, which encompass the predominant forms of object dynamics in real world environments.\label{scene_edit_type}

During task execution, the robot can only access current RGB-D data and poses from the environment.
Consequently, the robot must rely on single egocentric observations to actively perform change detection and determine whether objects in the scene have changed.
Upon detecting scene alterations, the robot is further required to autonomously collect data to update the 3DGS scene representation.

\subsection{GS-LTS System}
To address the aforementioned challenges, we propose GS-LTS, a 3D\textbf{GS}-based system tailored for \textbf{L}ong-\textbf{T}erm \textbf{S}ervice robots operating in indoor environments.
The system integrates four key modules as shown in Fig. \ref{fig:main}.
In the following, we provide an overview of the system’s operational workflow, with details of each module elaborated in Sec. \ref{Methodology}.

\textbf{Gaussian Mapping Engine}.
This module is tasked with generating the 3DGS scene representation for the robot during the system’s initialization phase.
Given a set of multi-view RGB images, depth maps, and their corresponding camera poses, the module trains a 3DGS model to effectively capture the scene’s geometric and visual characteristics.
In addition, to incorporate semantic information, we leverage the Segment Anything Model (SAM) \cite{sam} and open-vocabulary vision-language models (e.g., CLIP~\cite{clip}) to embed semantic features into the 3DGS representation.

\textbf{Multi-Task Executor}.
This module serves as an interface between the dense 3DGS representation and downstream tasks, enabling the robot to leverage the high-fidelity scene information encoded in the 3DGS for task planning and execution.
For instance, by matching textual features with 3D Gaussian semantic features, this module facilitates 3D localization for arbitrary text queries.
Additionally, it adapts the 3DGS into a 2D semantic map format to support existing methods such as object navigation. 

\textbf{Change Detection Unit}.
Since the robot must autonomously detect scene changes relying solely on single egocentric observations, we develop the Change Detection Unit, a lightweight module designed for long-term standby running in parallel with other downstream embodied tasks.
This module compares the robot’s current RGB observation with a rendered image generated from the 3DGS at the corresponding pose.
By employing a dual-branch strategy that analyzes both pixel-level differences and feature-level differences, the module effectively identifies the type and location of changes between the two compared frames.

\textbf{Active Scene Updater}.
Upon detecting scene changes and their locations, the Active Scene Updater module autonomously collects data and updates the long-term scene representation.
Initially, the robot follows a rule-based heuristic policy to navigate around the scene change region, capturing multi-view images.
Next, it applies 3DGS editing strategies to edit the target region.
Finally, the 3DGS representation is refined by fine-tuning with collected images.
This module enables GS-LTS to perform scene updates efficiently with minimal computational overhead.

\section{Methodology}
\label{Methodology}
In this section, we introduce the implementation and technical details of the GS-LTS system.

\subsection{3DGS Mapping Engine}
In this module, we employ semantic-aware 3D Gaussian Splatting (3DGS) for scene reconstruction. 3DGS provides an explicit scene representation through anisotropic Gaussians characterized by center position $\mu \in \mathbb{R}^3$, covariance matrix $\Sigma \in \mathbb{R}^{3\times3}$, opacity $o \in \mathbb{R}$, and color $c \in SH$ represented by spherical harmonics.
Through differentiable rendering, 3DGS synthesizes pixel colors via alpha blending:
\begin{equation}
    C = \sum_{i=1}^N c_i \alpha_i \prod_{j=1}^{i-1}(1-\alpha_j),
\end{equation}
where $i\ge2$ and $\alpha_i$ depends on $o_i$ and the projected 2D Gaussian's contribution to the current pixel.

We refer \cite{langsplat} to extend 3DGS for semantic fields. First, construct the vanilla 3DGS scene representation. Second, embed semantic features by leveraging SAM~\cite{sam} to generate multi-level masks $\{M^l\}_{l=1}^3$ and extract high-dimensional CLIP~\cite{clip} features $F^l = \mathrm{CLIP}(I \odot M^l) \in \mathbb{R}^{D}$. An autoencoder is used to compresses these features into low-dimensional semantic features $S^l=\mathrm{Encoder}(F^l)\in\mathbb{R}^d$ while preserving semantics through reconstruction regularization $\hat{F}^l = \mathrm{Decoder}(S^l)$. The low-dimensional semantic features are used to supervise the Gaussians' semantic attribute $s^l\in\mathbb{R}^d$ using view-consistent alpha blending:
\begin{equation}
    \hat{S}^l = \sum_{i=1}^N s^l_i \alpha_i \prod_{j=1}^{i-1}(1-\alpha_j).
\end{equation}

Thus, we obtain the semantic-aware 3DGS representation $\mathcal{G}=\{g_i\}_{i=1}^N$ which enables explicit 3D semantic representation while maintaining real-time rendering capabilities through the Gaussian representation.

\subsection{GS Multi-Task Executor}
Below, we illustrate how the 3DGS representation can be applied to robotic tasks, including 3D localization and object navigation, which are critical capabilities for numerous applications and serve to assess the geometric and semantic accuracy of 3DGS scene representation.


\subsubsection{Semantic Querying and Localization}
For an arbitrary text query, the relevance score $r(\phi_{\text{qry}},\phi_{\mathit{g}_i})$ between the CLIP embedding $\phi_{\text{qry}}$ and the semantic feature $\phi_{\mathit{g}_i} = \mathrm{Decoder}(s^l_i)$ of each 3D Gaussian is defined as:
\begin{equation}
    \min _j \frac{\exp \left(\phi_{\mathit{g}_i} \cdot \phi_{\text{qry}}\right)}
    {\exp \left(\phi_{\mathit{g}_i} \cdot \phi_{\text{qry}}\right)+\exp \left(\phi_{\mathit{g}_i} \cdot \phi_{\text{canon}}^j\right)},
\end{equation}
where $\phi_{\text{canon}}^{j}$ is the CLIP embedding of a predefined set of canonical phrases, including ``object'', ``things'', ``stuff'', and ``texture''.
The localization of the query is achieved by calculating the bounding box of matched Gaussians $\{\mathit{g}_i \mid r(\phi_{\text{qry}},\phi_{\mathit{g}_i}) > \tau_{\text{sim}}\}$, where $\tau_{\text{sim}}$ is a predefined threshold.
Due to the large number of 3D Gaussians in the scene, sparse sampling is applied in practice to perform semantic querying.

\subsubsection{2D Semantic Mapping and Navigation}
The 3DGS representation can be seamlessly converted into a 2D semantic map, ensuring compatibility with existing navigation and path planning methods.
The 2D semantic map is represented as a  $(L + 2) \times M \times M$ matrix,  where $M \times M$ represents the map size,
$L$ is the number of semantic categories, and the additional layers represent obstacles and explored area.
For $L$ navigation-relevant categories, we assign each 3D Gaussian the category with the highest relevance score.
The resulting semantic point cloud is voxelized and flattened along the Z-axis to form a 2D semantic map, enabling path planning and navigation to target categories.



\subsection{Change Detection Unit}
As mentioned in Sec.~\ref{scene_edit_type}, this module is designed to: (1) detect and classify three types of scene changes; (2) localize the target position $p_\text{world}$ in world coordinates.


\subsubsection{3DGS-based Change Detection}

The most intuitive method for detecting scene change is to analyze the discrepancies between the real-world image and the 3DGS rendered image, although 3DGS enables photo-realistic image rendering, there often exist pixel-wise errors with the real-world observation, making the direct computation of absolute pixel differences between the two images yield sub-optimal results. 

Therefore, following the practice of~\cite{lu20253dgs}, we employ a dual-branch strategy for scene change detection.
As shown in Fig. \ref{fig:main}, given the the real-world camera captured image $I_{\text{real}}$ and 3DGS rendered image $I_{GS}$ from the current viewpoint, we first calculate the sum of absolute pixel differences across all 3 channels between the two images, which is truncated via threshold $\tau_{\text{GS}}$ to obtain the pixel-level binary mask:
\begin{equation}
M_{\text{pixel}} = \left(\sum\nolimits_{\text{chn}=1}^3 \left| I_{\text{real},\text{chn}} - I_{\text{GS},\text{chn}} \right| > \tau_{\text{GS}}\right).
\end{equation}

Next, we compute normalized EfficientSAM~\cite{effsam} feature maps $I_{\text{real}, \text{SAM}}$ and $I_{\text{GS}, \text{SAM}}$ that robustly represent significant regions, then calculate their cosine similarity truncated by $\tau_{\text{feat}}$ to obtain the feature-level binary mask:
\begin{equation}
M_{\text{feat}} = \left(\left\langle I_{\text{GS}, \text{SAM}}, I_{\text{real}, \text{SAM}} \right\rangle > \tau_{\text{feat}}\right).
\end{equation}

Finally, the combined binary mask can be obtained using pixel-by-pixel multiplication of the dual-branch binary masks $M_{\text{comb}}=M_{\text{pixel}} \odot M_{\text {feat}}$.
We hypothesize that when the total area of $M_\text{comb}$ exceeds the threshold $\tau_{\text{change}}$, a scene change occurs, thereby triggering scene change prediction.

\subsubsection{Scene Change Prediction}
We posit that all potential change regions reside within $M_\text{comp}$, where noise areas constitute a small portion. Therefore, we first extract connected components $\{R_i\}_{i=1}^N$ from $M_\text{comp}$, sorted in descending order by their area. Based on the distinct change types, we hypothesize that: (i) relocation operations are geometrically constrained to the first two largest connected components ($R_1$ and $R_2$), while (ii) addition/removal operations manifest exclusively within the dominant connected components ($R_1$). We first formulate the following dual-region matching criterion to identify \textbf{relocation} operation:\label{sec:SceneChangePrediction}
\begin{equation}
    \Gamma_{\text{match}}= 
    \begin{aligned}[t]
        &\underbrace{\frac{\min(A(R_1), A(R_2))}{\max(A(R_1), A(R_2))}}_{\text{Area similarity}} > \tau_a \\
        &\wedge \quad 
        \underbrace{\|\eta(R_1)-\eta(R_2)\|_2}_{\text{Spatial distance}} < \tau_d,
    \end{aligned}
    \label{eq:relocation_condition}
\end{equation}
where $A(\cdot)$ denotes region area, $\eta(\cdot)$ for centroid coordinates. 

Then, for the largest connected components $R_{1}$, we compute its centroid $p_c=(u_c,v_c)$, and  sample depth from real-world depth map $D_{\text{cam}}$ and 3DGS rendered depth map $D_{\text{GS}}$:
\begin{equation}
    \begin{cases}
        d_{\text{real}} = D_{\text{real}}(p_c) \\
        d_{\text{GS}} = D_{\text{GS}}(p_c)
    \end{cases}
    \label{eq:depth_sampling}
\end{equation}

The change type is determined through depth difference:
\begin{equation}
    \Delta d = d_{\text{real}} - d_{\text{GS}} \Rightarrow 
    \text{Type} = 
    \begin{cases} 
        \text{Addition}, & \Delta d<-\epsilon \\
        \text{Removal}, & \Delta d>\epsilon \\
        \text{Unchanged}, & |\Delta d|\leq\epsilon
    \end{cases}
    \label{eq:depth_decision}
\end{equation}

To obtain $p_\text{world}$, here we construct a pseudo depth map $D_{\text{pseudo}}(u,v) = \min(D_{\text{real}}(u,v),D_{\text{GS}}(u,v))$, then generate camera-coordinate key point $p_{\text{cam}} \in \mathbb{R}^3$ based on event type:
\begin{equation}
    p_{\text{cam}} = 
    \begin{cases}
        [u_c, v_c, D_{\text{pseudo}}(u_c, v_c)]^\top, &  
        \begin{aligned}
            &\text{Addition}/\\[-3.7pt]
            &\text{Removal}
        \end{aligned} \\[6pt]
        \begin{aligned}
            \frac{1}{2}\big(&[c(R_1); D_{\text{pseudo}}(c(R_1))] \\
            + &[c(R_2); D_{\text{pseudo}}(c(R_2))]\big)^\top
        \end{aligned},
        & \text{Relocation}
    \end{cases}
    \label{eq:keypoint_generation}
\end{equation}

Then, $p_{\text{world}}$ can  be calculated using camera-to-world transformation matrix $T_{\text{cam}}^{\text{world}}$:  
\begin{equation}
    p_{\text{world}} = T_{\text{cam}}^{\text{world}} \cdot \begin{bmatrix}p_{\text{cam}} \\ 1\end{bmatrix}.
    \label{eq:p_c2w}
\end{equation}




\subsection{Active Scene Updater}

\subsubsection{Active Data Collection}

We design a rule-based heuristic policy to enable the robot to autonomously capture images of scene change regions.
The core of this policy involves positioning the robot to face the target region, with the region as the circle’s center, and moving left or right along the tangential direction.
The robot first adjusts its distance from the target region to ensure a successful tangential movement.
After each movement, the robot reorients its viewpoint toward the target and readjusts its distance to capture images.
Following this policy, the robot first moves $K/2$ steps to the left, then $K$ steps to the right, collecting images $\{I_{\text{real},i}\}_{i=1}^K$, depth maps $\{D_{\text{real},i}\}_{i=1}^K$ and camera poses $\{T_i\}_{i=1}^K$ from $K$ viewpoints of the scene change region during the process. Next, the robot obtains combined masks $\{M_{\text{comb},k}\}_{k=1}^K$ using the method in Sec.~\ref{sec:SceneChangePrediction}.


\subsubsection{Gaussian Editing based Scene Update}
We adopt a pre-editing and fine-tuning strategy to achieve scene update. Distinct pre-editing protocols are implemented for different scene change types: (i) For Addition, we directly instantiate new Gaussians in target region; (ii) For Removal, we first localize target objects and then prune the associated Gaussians followed by scene inpainting to fix the hole; (iii) For Relocation, we execute a delete-then-add strategy that removes the associated Gaussians and then instantiate new Gaussians in target region.
To identify the associated Gaussians, we refer to \cite{chen2024gaussianeditor} for defining a voting function $\mathcal{V}$ that localizes target Gaussians within $\{M_{\text{comb},k}\}_{k=1}^K$,
\begin{equation}
    \mathcal{V}(g_i) = \sum_{k=1}^K \mathbb{I}\left[\pi(T_k \mu_i) \in \{M_{\text{comb},k}\}_{k=1}^K \right],
    \label{eq:voting}
\end{equation}
where $\pi(\cdot)$ is the projection function, the target Gaussians are selected based on the proportion $\rho$ of its presence within the mask region: $\mathcal{G}_{\text{target}} = \{g_i | \mathcal{V}(g_i) > \rho  \cdot K\}.$

The workflow of our pre-editing method is as follows:

For \textbf{addition}:
\begin{enumerate}[label=\roman*.]
    \item Generate object points $\mathcal{P}_{\text{add}} = \{p_j\}$ via depth map $\{D_{\text{real},k}\odot M_{\text{comb},k}\}_{k=1}^K$.
    \item Pass new semantic feature $s_{\text{manual}}$.
    \item Find nearest neighbors and inherit attributes:
    \begin{enumerate}
        \item For each $p_j \in \mathcal{P}_{\text{add}}$: $g_{\text{nn}}^j = \mathop{\arg\min}\limits_{g_m \in \mathcal{G}} \|\mu_m - p_j\|_2$,
        \item Extract covariance matrix $\Sigma_{\text{nn}}^j$, opacity $o_{\text{nn}}^j$ and color $c_{\text{nn}}^j$ from $g_{\text{nn}}^j$.
    \end{enumerate} 
    \item Create new Gaussians $\mathcal{G}_{\text{add}}$ and insert to $\mathcal{G}$:
    \begin{equation} 
    \mathcal{G}_{\text{add}} = \left\{ g_i \,\middle|\, 
        (p_j, \Sigma_{\text{nn}}^j, o_{\text{nn}}^j, c_{\text{nn}}^j, s_{\text{manual}} )
    \right\}.
    \end{equation}
\end{enumerate}

For \textbf{removal}:
\begin{enumerate}[label=\roman*.]
    \item Delete $\mathcal{G}_{\text{target}}$ from $\mathcal{G}$.
    \item Generate hole inpainting points $\mathcal{P}_{\text{fill}} = \{p_j\}$ via depth map $\{D_{\text{real},k}\odot M_{\text{comb},k}\}_{k=1}^K$.
    \item Find nearest neighbors and inherit attributes:
    \begin{enumerate} 
        \item For each $p_j \in \mathcal{P}_{\text{fill}}$: $g_{\text{nn}}^j = \mathop{\arg\min}\limits_{g_m \in \mathcal{G}} \|\mu_m - p_j\|_2$,
        \item Extract covariance matrix $\Sigma_{\text{nn}}^j$, opacity $o_{\text{nn}}^j$, color $c_{\text{nn}}^j$ and semantic  $s_{\text{nn}}^j$ from $g_{\text{nn}}^j$.
    \end{enumerate}
    \item Create new Gaussians $\mathcal{G}_{\text{add}}$ and insert into $\mathcal{G}$: 
    \begin{equation} 
    \mathcal{G}_{\text{add}} = \left\{ g_i \,\middle|\, 
            (p_j, \Sigma_{\text{nn}}^j, o_{\text{nn}}^j, 
            c_{\text{nn}}^j, s_\text{nn}^j)
    \right\}.
    \end{equation}
\end{enumerate}

For \textbf{relocation}:
\begin{enumerate}[label=\roman*.]
    \item Obtain $\mathcal{G}_{\text{target}}$ and extract colors and semantics $\mathcal{C}, \mathcal{S} = \{c_j, s_j | g_j \in \mathcal{G}_{\text{target}}\}$.
    \item Delete $\mathcal{G}_{\text{target}}$ and generate target points $\mathcal{P}_{\text{dest}}$ via depth map $\{D_{\text{real},k}\odot M_{\text{comb},k}\}_{k=1}^K$ at destination.
    \item For each $p_j \in \mathcal{P}_{\text{dest}}$, $g_{\text{nn}}^j = \mathop{\arg\min}\limits_{g_m \in \mathcal{G}} \|\mu_m - p_j\|_2$, create new Gaussians $\mathcal{G}_{\text{add}}$ and insert into $\mathcal{G}$:
    \begin{equation} 
    \mathcal{G}_{\text{add}} = \left\{ g_i \,\middle|\, 
            (p_i, \Sigma_{\text{nn}}^j, o_{\text{nn}}^j, 
            c_{\text{nn}}\sim\mathcal{C}, s_{\text{nn}}\sim\mathcal{S})
    \right\}.
    \end{equation} 
    \item Hole inpainting to source region as aforementioned.
    
\end{enumerate}


Finally, we perform post-training to further fine-tune the Gaussians and refine the reconstruct quality.
\begin{table*}
    \caption{Scene Update Quality Evaluated by Image Rendering Metrics (250 Fine-Tuning Iterations).}
    \vspace{-0.8em}
    \centering
    \begin{tabular}{*{14}{c}}
        \toprule
        \multirow{2}{*}{Method} & \multirow{2}{*}{View} & \multicolumn{3}{c}{addition} & \multicolumn{3}{c}{relocation} & \multicolumn{3}{c}{removal} & \multicolumn{3}{c}{overall} \\
        \cmidrule(lr){3-5} \cmidrule(lr){6-8} \cmidrule(lr){9-11} \cmidrule(lr){12-14}
        & & SSIM$\uparrow$ & PSNR$\uparrow$ & LPIPS$\downarrow$ & SSIM$\uparrow$ & PSNR$\uparrow$ & LPIPS$\downarrow$ & SSIM$\uparrow$ & PSNR$\uparrow$ & LPIPS$\downarrow$ & SSIM$\uparrow$ & PSNR$\uparrow$ & LPIPS$\downarrow$ \\
        \midrule
        \multirow{2}{*}{Baseline} & near & 0.87 & 24.74 & 0.24 & 0.89 & 25.94 & 0.22 & 0.91 & 29.79 & 0.20 & 0.89 & 27.20 & 0.22 \\
                                   & far  & 0.88 & 25.52 & 0.20 & 0.89 & 26.23 & 0.19 & 0.91 & 28.64 & 0.17 & 0.89 & 27.04 & 0.19 \\
        \midrule
        \multirow{2}{*}{GS-LTS} & near & \textbf{0.88} & \textbf{26.60} & \textbf{0.22} & \textbf{0.91} & \textbf{28.58} & \textbf{0.19} & \textbf{0.93} & \textbf{32.08} & \textbf{0.17} & \textbf{0.91} & \textbf{29.40} & \textbf{0.19} \\
                                  & far  & \textbf{0.89} & \textbf{26.93} & \textbf{0.18} & \textbf{0.90} & \textbf{28.04} & \textbf{0.17} & \textbf{0.92} & \textbf{30.55} & \textbf{0.15} & \textbf{0.90} & \textbf{28.74} & \textbf{0.16} \\
        \bottomrule
    \end{tabular}
    \vspace{-1.8em}
    \label{table:Scene Change Detection}
\end{table*}


\section{Experiments}
This section details a comprehensive evaluation of GS-LTS across simulation and real-world settings.

\subsection{Scene Change Adaptation Benchmark}

\subsubsection{Settings}
To assess the robot's ability to adapt to scene changes in dynamic environments, we present a novel Scene Change Adaptation Benchmark constructed on the AI2-THOR simulation platform \cite{ai2thor}. AI2-THOR offers a comprehensive suite of APIs such as \textit{InitialRandomSpawn}, \textit{DisableObject}, and \textit{PlaceObjectAtPoint} that facilitate direct manipulation of scene objects, which we exploit to design three distinct types of scene update tasks: \textbf{relocation}, \textbf{addition}, and \textbf{removal} of objects.
\begin{figure}[t]
\centering
\includegraphics[width=1\linewidth,height=0.4\linewidth]{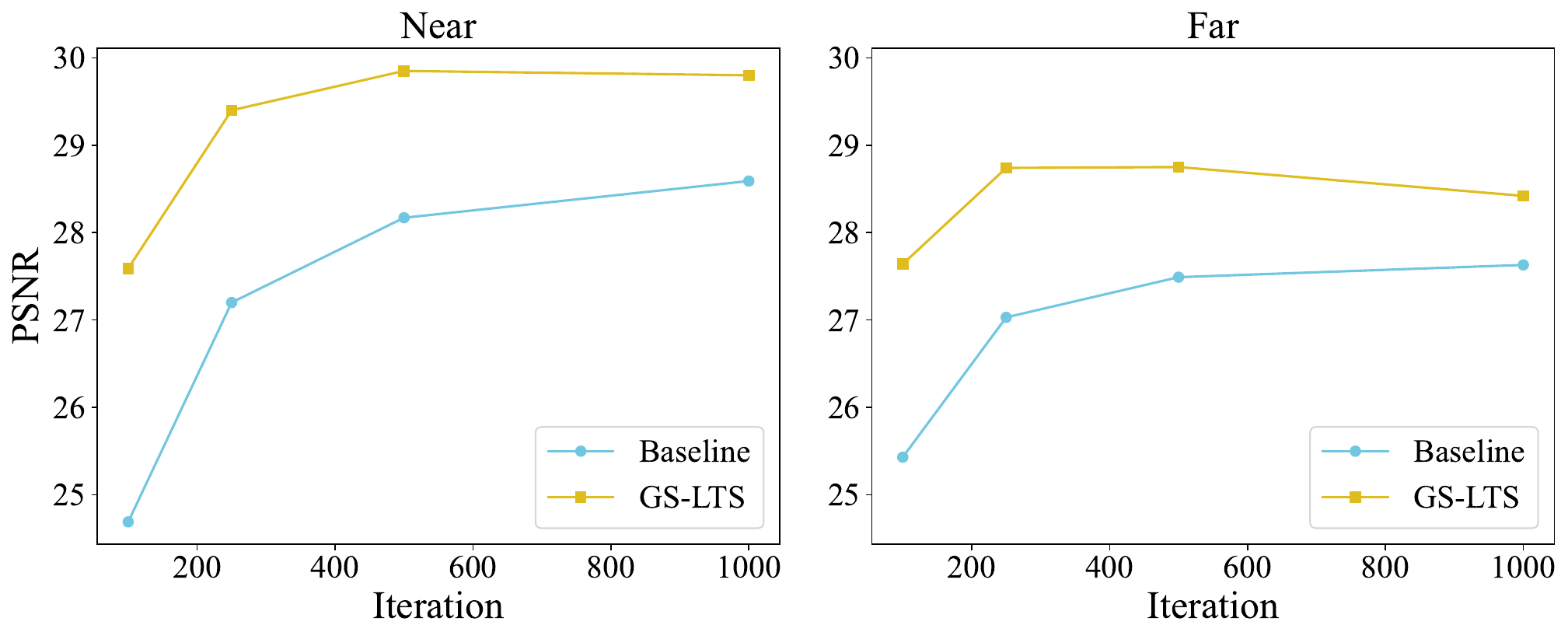}
\vspace{-1.8em}
\caption{Impact of fine-tuning iterations on scene update quality.}
\label{fig:psnr_line}
\vspace{-1.6em}
\end{figure}
The benchmark is generated by automatically traversing combinations of editable objects, containers, and placement positions, which enables the sampling of an extensive range of scene changes.
Each scene change task is recorded via a configuration script containing environment metadata (e.g., initial viewpoint) and a sequential action list specifying the operations to transform objects from their default state to the updated state.
Additionally, each task involves 20 test viewpoints capturing the scene change region (10 near-range and 10 far-range).
As changed objects typically occupy a small fraction of the field of view, we generate test images by expanding ground-truth change bounding boxes by 50 pixels in all directions. Scene update quality is then assessed using PSNR, SSIM, and LPIPS metrics.


For evaluation, the robot is initialized at the starting pose of each scene change task.
The Change Detection Unit is first executed to generate predictions, after which we assess whether the predicted scene change type matches the ground-truth type and whether the prediction error of the scene change region is within 1 meter.
For tasks with accurate predictions, active data collection and Gaussian editing-based scene update are performed.
During scene representation updates, GS-LTS first employs pre-editing method, followed by fine-tuning of 3DGS to refine object geometry and visual details.
In contrast, the baseline method directly fine-tunes 3DGS using multi-view data collected by GS-LTS.

In this experiment, we sample 459 scene change tasks, achieving a 74.5\% accuracy in predicting change type and target region with GS-LTS.
Scene updates are tested on 342 tasks, with results after 250 fine-tuning iterations reported in Table \ref{table:Scene Change Detection}.
Experimental results demonstrate that our method achieves superior performance for both type of viewpoints, outperforming the baseline across all metrics.

\begin{figure}[t]
\centering
\includegraphics[width=1\linewidth,height=0.55\linewidth]{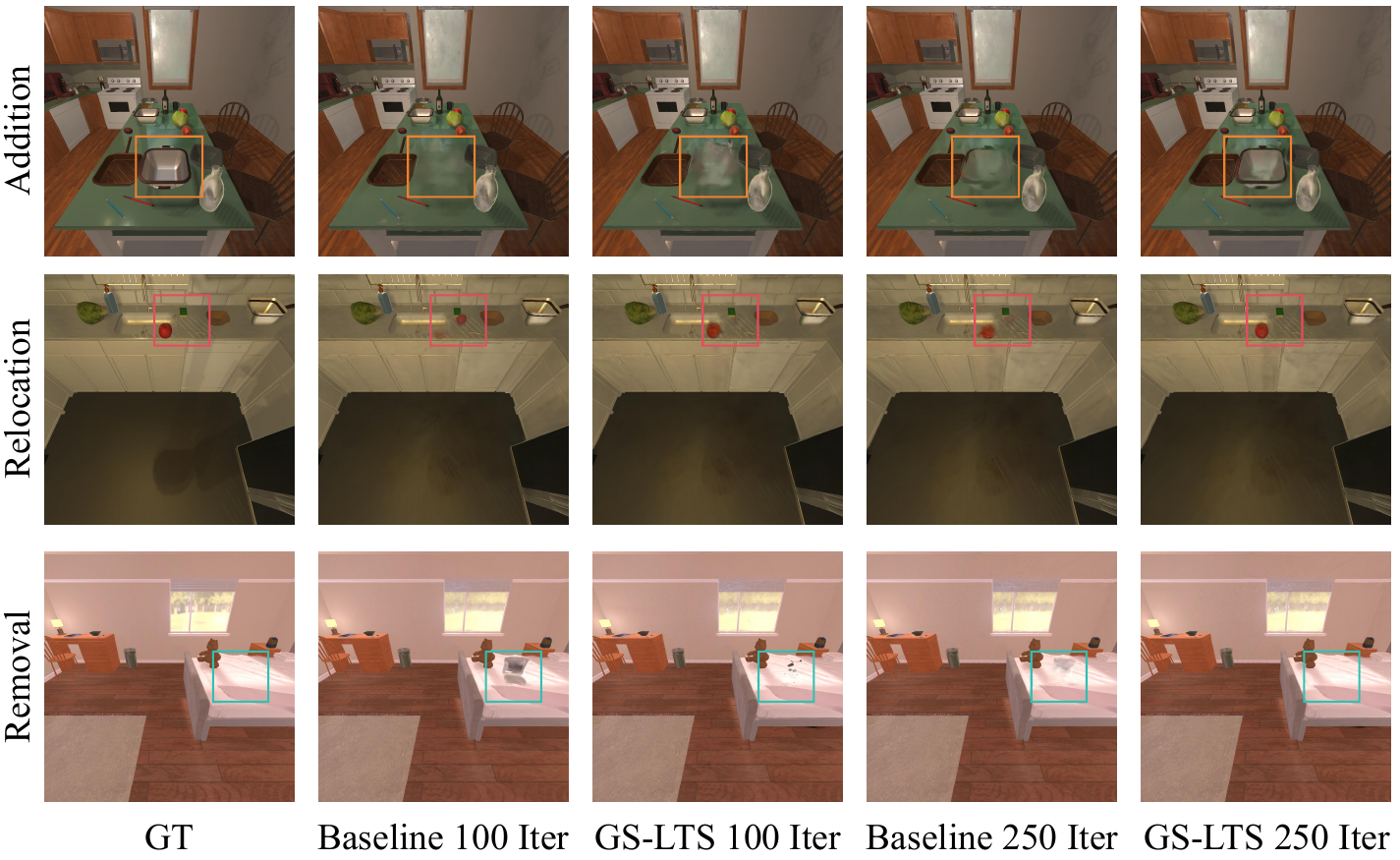}
\vspace{-1.8em}
\caption{Rendering results after different fine-tuning iterations.}
\label{fig:scene_change}
\vspace{-0.4em}
\end{figure}

Additionally, as shown in Fig. \ref{fig:psnr_line}, we evaluate various fine-tuning iterations and statistically analyze the overall PSNR metrics for both type of viewpoints. The results demonstrate that our approach consistently outperforms the baseline across all settings. Notably, GS-LTS achieves superior scene update quality with fewer fine-tuning iterations, highlighting its ability to deliver efficient, low-cost scene updates.
Fig. \ref{fig:scene_change} presents the quantitative results of GS-LTS, showcasing one representative case from each of three scene changes.
The rendering results more intuitively demonstrate that GS-LTS achieves superior and faster scene update capabilities, while the baseline method requires significantly more iterations to obtain comparable outcomes.

\begin{table}[t]
    \centering
    \caption{3D Localization Results (Bottom Part for Ablation Study).}
    \vspace{-0.8em}
    \begin{tabular}{lccc}
      \toprule
      Feature Source & mIoU & Acc (IoU$>$0.5) & Acc (IoU$>$0.3) \\
      \midrule
      CLIP & 40.6 & 42.9 & 52.1 \\
      GT & \textbf{60.9} & \textbf{73.6} & \textbf{81.8} \\
    \midrule
      CLIP (300$\times$300 res)  & 24.6 & 29.0 & 30.7 \\
     CLIP (8 dim)  & 32.2 & 35.7 & 43.3 \\
      CLIP (60\% data)   & 36.6 & 41.1 & 46.1 \\
      \bottomrule
    \end{tabular}
    \label{table:3DLocalization}
    \vspace{-1.8em}
\end{table}

\begin{figure}[t]
\centering
\includegraphics[width=1\linewidth,height=0.20\linewidth]{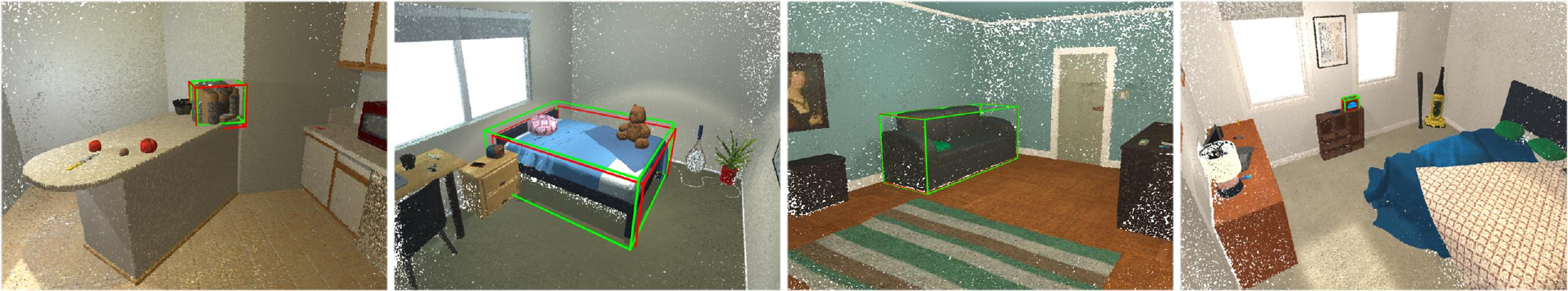}
\vspace{-1.8em}
\caption{\textbf{3D Localization Examples.} Red bounding boxes indicate the results from GS-LTS (GT), while green ones from GS-LTS (CLIP).}
\label{fig:3dbbox}
\vspace{-0.6em}
\end{figure}

\begin{table}[t]\scriptsize
    \centering
    \caption{Object Navigation Results Across Three Room Types.}
    \vspace{-0.8em}
    \resizebox{\linewidth}{!}{
    \begin{tabular}{lcccccccc}
      \toprule
      \multirow{2}{*}{Method} & \multicolumn{2}{c}{kitchen} & \multicolumn{2}{c}{living room} & \multicolumn{2}{c}{bedroom} & \multicolumn{2}{c}{overall} \\
      \cmidrule(lr){2-3}\cmidrule(lr){4-5}\cmidrule(lr){6-7}\cmidrule(lr){8-9}
       & SPL & SR  & SPL & SR  & SPL & SR  & SPL & SR
       \\
      \midrule
      SAVN \cite{wortsman2019learning} & 17.8 & 43.6 & 7.7 & 21.6 & 8.7 & 29.2 & 11.4 & 31.5 \\
      GVE \cite{moghaddam2021optimistic}  & 17.9 & 45.6 & 9.4 & 25.2 & 8.1 & 27.6 & 11.8 & 32.8  \\
      HZG \cite{he2024relation}  & 48.7 & 74.6 & 41.5 & 60.7 & 32.2 & \textbf{59.1} & 40.8 & 64.8  \\
      GS-LTS (CLIP)  & 45.7 & 59.0 & 40.0 & 52.2 & 41.5 & 54.6 & 42.2 & 55.3  \\
      GS-LTS (GT)  & \textbf{64.0} & \textbf{86.4} & \textbf{68.4} & \textbf{90.3} & \textbf{44.8} & 56.4 & \textbf{59.1} & \textbf{77.7}  \\
      \bottomrule
    \end{tabular}}
    \label{table:Object-Navigation}
    \vspace{-2.8em}
\end{table}

\subsection{Multi-Task Experiment}
To assess the geometry and semantic fidelity of GS-LTS, we conduct experiments on 3D localization and object navigation.
We evaluate two 3DGS representations embedding ground-truth semantics and CLIP semantics, denoted as GS-LTS (GT) and GS-LTS (CLIP), respectively.

\subsubsection{3D Localization}
For the 3D localization task, semantic quality is quantitatively assessed by calculating the Intersection over Union (IoU) of the 3D bounding boxes.
A 3D bounding box is deemed accurately localized if its IoU with the ground-truth bounding box exceeds a predefined threshold.
Based on this criterion, we compute the Acc (IoU$>$threshold) metric to evaluate localization accuracy. We evaluate localization performance across 12 different object categories, including \textit{AlarmClock, ArmChair, Bed, Bread, Chair, CoffeeMachine, DeskLamp, DiningTable, Dresser, Dumbbell, RemoteControl and Sofa}.


As shown in the top part of Table \ref{table:3DLocalization}, GS-LTS (GT) significantly outperforms GS-LTS (CLIP) in terms of both mIoU and accuracy metrics, highlighting the critical importance of precise semantic cues.
Fig. \ref{fig:3dbbox} presents qualitative results for the 3D localization task.
The 3D bounding boxes generated by GS-LTS (GT) generally exhibit a closer alignment with the objects compared to those generated by GS-LTS (CLIP).
These precise bounding boxes further highlight the advantages and potential of employing 3DGS as a scene representation.

\subsubsection{Object Navigation}
For the object navigation task, we adopt the experimental protocol proposed by SAVN \cite{wortsman2019learning}, with a modification to limit the evaluation to three scene types: kitchens, living rooms, and bedrooms. Bathrooms are excluded due to their constrained spatial scale and simplistic layouts.
Performance is assessed using the Success weighted by Path Length (SPL) and Success Rate (SR).

\begin{figure}[t]
\centering
\includegraphics[width=1\linewidth,height=0.65\linewidth]{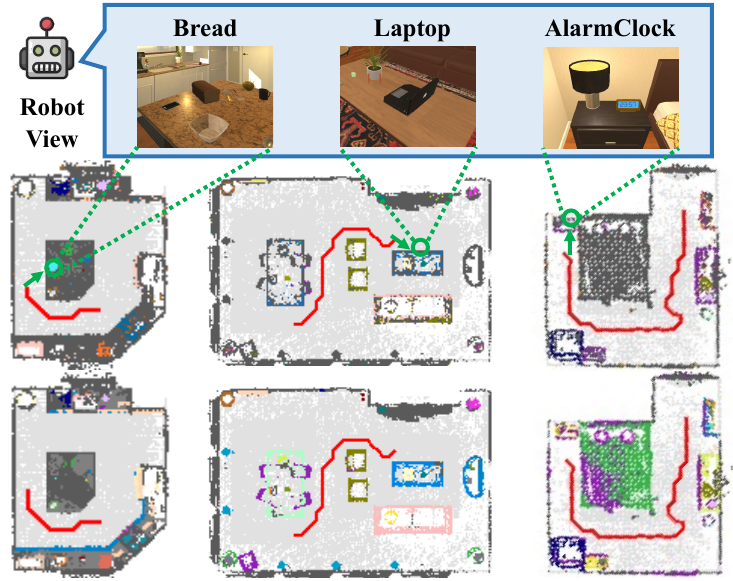}
\vspace{-1.8em}
\caption{\textbf{Object Navigation Examples.} The mid row displays semantic maps and navigation trajectories generated by GS-LTS (CLIP), the bottom row illustrates the corresponding outputs of GS-LTS (GT).}
\label{fig:nav}
\vspace{-0.8em}
\end{figure}

We compare GS-LTS with classical methods.
Notably, these classical methods trained on the AI2THOR involve exploration and navigation within a single episode.
In contrast, GS-LTS leverages prebuilt 3DGS representations and performs training-free navigation directly from a 2D semantic map, employing a deterministic policy (Fast Marching Method).
This experimental setup is designed to validate the feasibility of 3DGS-based robotic navigation using existing benchmark.
As shown in Table \ref{table:Object-Navigation}, GS-LTS (GT) outperforms other approaches across most metrics, while GS-LTS (CLIP) also demonstrates competitive performance, particularly on the SPL metric.
Semantic maps and trajectories for three example navigation tasks are illustrated in Fig. \ref{fig:nav}.

\subsection{Ablation Study}



To examine the effect of initial training data on 3DGS representations, we perform an ablation study on the 3D localization task, with results reported in the bottom part of Table \ref{table:3DLocalization}.
We analyze how reduced image resolution, feature dimension and data volume affect GS-LTS (CLIP) performance.
Experiments reveal that lowering resolution from 1,000$\times$1,000 to 300$\times$300 decreases mIoU by 16.0\% and significantly reduces accuracy.
Decreasing the feature dimension from 32 to 8 results in a performance drop of mIoU from 40.6\% to 32.2\%, indicating that lower-dimensional representations degrade the quality of the learned latent space, as convergence of autoencoders becomes more challenging with 8-dimensional features. 
Reducing the training dataset from an average of 670 images to 60\% of the data lowers mIoU by 4.0\%.
Notably, CLIP features computed from SAM-segmented masks rely heavily on high-resolution images for small object recognition.
While smaller data volume affect fine details, the impact is minimal for objects visible from multiple viewpoints, such that overall performance decline remains limited.


\begin{figure}[t]
\centering
\includegraphics[width=1\linewidth,height=0.68\linewidth]{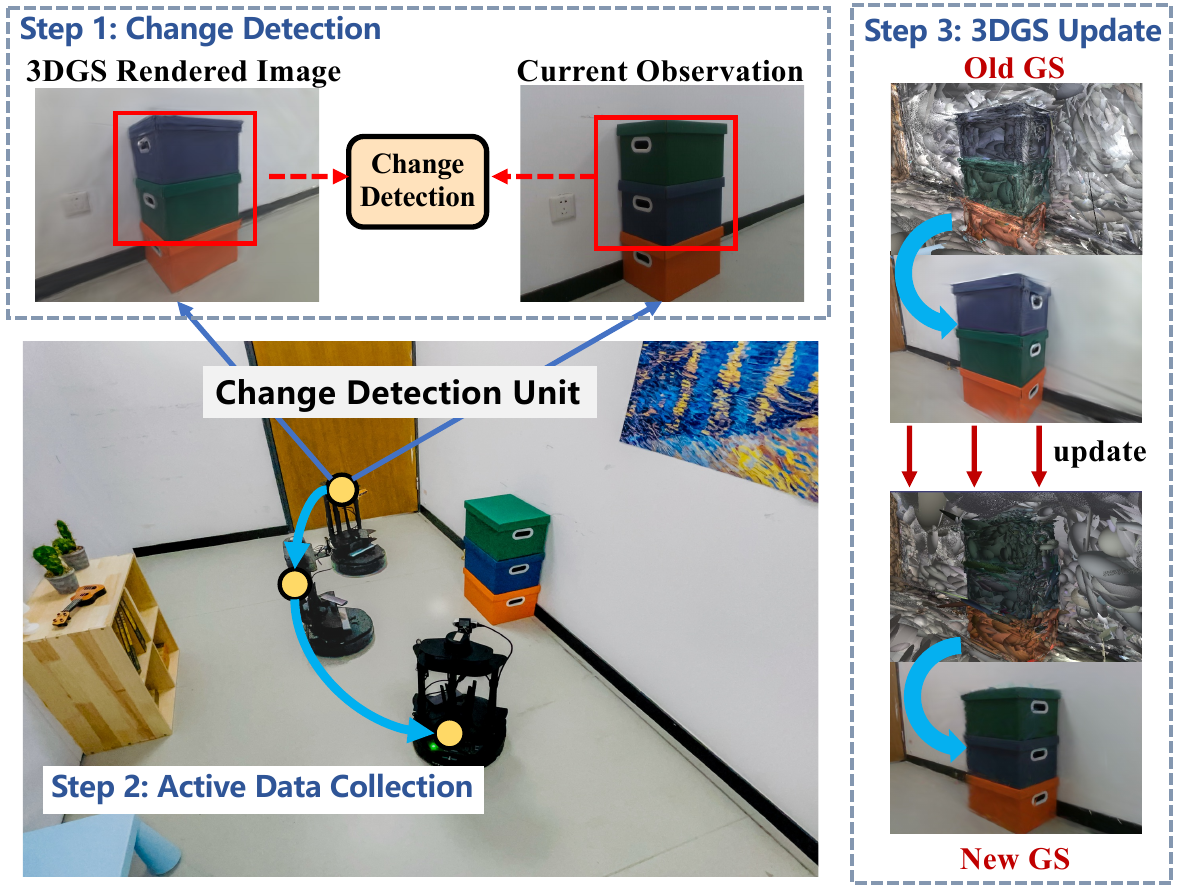}
\vspace{-1.8em}
\caption{Real robot performing scene change adaptation.}
\label{fig:real}
\vspace{-1.8em}
\end{figure}

\subsection{Application in Real-world Robot System}
To demonstrate the real-world applicability of the GS-LTS system, we conducted experiments with a real robot.
We utilize a Microsoft Azure Kinect DK camera to scan a pre-arranged room, capturing data to train a 3DGS representation of the scene.
Unlike simulation environments, where precise robot poses can be obtained directly from an environment API, such information is unavailable in real-world settings.
To address this, we augment the GS-LTS system with a relocalization module tailored for real-world operation.
Here, we first obtain a coarse pose estimation through ORB visual feature matching, then employ iComMa \cite{icomma} to perform pose refinement to obtain an optimized precise pose estimation.
To assess the robot's ability to adapt to scene changes, we reposition three stacked colored storage bins within the room.
As the robot approaches the vicinity of the bins, the Change Detection Unit identifies discrepancies in the current scene.
It then actively collects multi-view images to update 3DGS.
Fig. \ref{fig:real} illustrates the changes in the 3DGS representation and rendered images before and after the adaptation.
These results validate that the GS-LTS system can effectively operate in real-world environments and adapt to dynamic scene changes.
For a detailed experimental video, please refer to our website.

\section{Discussion}
\subsection{Resource  Overhead}
The entire system operates efficiently on a single NVIDIA GeForce RTX 4090 GPU.
The GS-LTS system completes vanilla 3DGS reconstruction in $\sim$15 minutes, with subsequent 32 dimensional Gaussian semantic learning requiring $\sim$1 hour. Our experiments show 250 training iterations achieve superior scene updates (0.91 SSIM / 29.07 PSNR) versus 1,000-iteration baselines, with $\le$10s training.

\subsection{Limitation and Future Work}

The GS-LTS system advances adaptive modeling for 3DGS-based robotic systems in long-term dynamic environments, yet several challenges remain before achieving widespread real-world deployment. Below, we discuss key limitations and promising directions for improvement.

First, efficient large-scale representation is a challenge for vanilla 3DGS, which struggles with expansive scenes like factories, requiring more storage-efficient solutions.

Second, robot control could be improved with learning-based policies to enhance adaptability in complex scenarios.

Finally, highly dynamic environments present an additional challenge. GS-LTS focuses on medium-term changes, not real-time dynamics like moving objects or human interactions. Future 3DGS-based dynamic reconstruction will enhance support for tasks like cooking or household assistance, improving more realistic long-term autonomy.


\section{Conclusions}

In this work, we introduce GS-LTS, a 3DGS-based system designed for long-term service robots operating in dynamic environments.
By integrating object-level change detection, multi-view observation, and efficient Gaussian editing-based scene updates, GS-LTS enables robots to adapt to scene variations over time. Additionally, we propose a scalable simulation benchmark for evaluating object-level scene changes, facilitating systematic assessment and sim-to-real transfer.
Experimental results demonstrate that GS-LTS achieves faster and higher-quality scene updates, advancing the applicability of 3DGS for long-term robotic operations.







\bibliographystyle{IEEEtran}
\bibliography{root}

\begin{thebibliography}{10}
\providecommand{\url}[1]{#1}
\csname url@rmstyle\endcsname
\providecommand{\newblock}{\relax}
\providecommand{\bibinfo}[2]{#2}
\providecommand\BIBentrySTDinterwordspacing{\spaceskip=0pt\relax}
\providecommand\BIBentryALTinterwordstretchfactor{4}
\providecommand\BIBentryALTinterwordspacing{\spaceskip=\fontdimen2\font plus
\BIBentryALTinterwordstretchfactor\fontdimen3\font minus
  \fontdimen4\font\relax}
\providecommand\BIBforeignlanguage[2]{{%
\expandafter\ifx\csname l@#1\endcsname\relax
\typeout{** WARNING: IEEEtran.bst: No hyphenation pattern has been}%
\typeout{** loaded for the language `#1'. Using the pattern for}%
\typeout{** the default language instead.}%
\else
\language=\csname l@#1\endcsname
\fi
#2}}

\bibitem{kerbl20233d}
B.~Kerbl, G.~Kopanas, T.~Leimk{\"u}hler, and G.~Drettakis, ``3d gaussian
  splatting for real-time radiance field rendering.'' \emph{ACM Trans. Graph.},
  vol.~42, no.~4, pp. 139--1, 2023.

\bibitem{keetha2024splatam}
N.~Keetha, J.~Karhade, K.~M. Jatavallabhula, G.~Yang, S.~Scherer, D.~Ramanan,
  and J.~Luiten, ``Splatam: Splat track \& map 3d gaussians for dense rgb-d
  slam,'' in \emph{CVPR}, 2024, pp. 21\,357--21\,366.

\bibitem{wu2024hgs}
K.~Wu, K.~Zhang, Z.~Zhang, M.~Tie, S.~Yuan, J.~Zhao, Z.~Gan, and W.~Ding,
  ``Hgs-mapping: Online dense mapping using hybrid gaussian representation in
  urban scenes,'' \emph{RAL}, 2024.

\bibitem{langsplat}
M.~Qin, W.~Li, J.~Zhou, H.~Wang, and H.~Pfister, ``Langsplat: 3d language
  gaussian splatting,'' in \emph{CVPR}, 2024, pp. 20\,051--20\,060.

\bibitem{zhu20243d}
S.~Zhu, G.~Wang, D.~Kong, and H.~Wang, ``3d gaussian splatting in robotics: A
  survey,'' \emph{arXiv preprint arXiv:2410.12262}, 2024.

\bibitem{wortsman2019learning}
M.~Wortsman, K.~Ehsani, M.~Rastegari, A.~Farhadi, and R.~Mottaghi, ``Learning
  to learn how to learn: Self-adaptive visual navigation using meta-learning,''
  in \emph{CVPR}, 2019, pp. 6750--6759.

\bibitem{chaplot2020object}
D.~S. Chaplot, D.~P. Gandhi, A.~Gupta, and R.~R. Salakhutdinov, ``Object goal
  navigation using goal-oriented semantic exploration,'' in \emph{NeurIPS},
  2020, pp. 4247--4258.

\bibitem{chen2019suma++}
X.~Chen, A.~Milioto, E.~Palazzolo, P.~Giguere, J.~Behley, and C.~Stachniss,
  ``Suma++: Efficient lidar-based semantic slam,'' in \emph{IROS}, 2019, pp.
  4530--4537.

\bibitem{mildenhall2021nerf}
B.~Mildenhall, P.~P. Srinivasan, M.~Tancik, J.~T. Barron, R.~Ramamoorthi, and
  R.~Ng, ``Nerf: Representing scenes as neural radiance fields for view
  synthesis,'' \emph{Communications of the ACM}, vol.~65, no.~1, pp. 99--106,
  2021.

\bibitem{jin2024gs}
R.~Jin, Y.~Gao, Y.~Wang, Y.~Wu, H.~Lu, C.~Xu, and F.~Gao, ``Gs-planner: A
  gaussian-splatting-based planning framework for active high-fidelity
  reconstruction,'' in \emph{IROS}.\hskip 1em plus 0.5em minus 0.4em\relax
  IEEE, 2024, pp. 11\,202--11\,209.

\bibitem{chen2024gaussianeditor}
Y.~Chen, Z.~Chen, C.~Zhang, F.~Wang, X.~Yang, Y.~Wang, Z.~Cai, L.~Yang, H.~Liu,
  and G.~Lin, ``Gaussianeditor: Swift and controllable 3d editing with gaussian
  splatting,'' in \emph{CVPR}, 2024, pp. 21\,476--21\,485.

\bibitem{kunze2018artificial}
L.~Kunze, N.~Hawes, T.~Duckett, M.~Hanheide, and T.~Krajn{\'\i}k, ``Artificial
  intelligence for long-term robot autonomy: A survey,'' \emph{RAL}, vol.~3,
  no.~4, pp. 4023--4030, 2018.

\bibitem{jones2012slocum}
C.~P. Jones, ``Slocum glider persistent oceanography,'' in \emph{AUV}.\hskip
  1em plus 0.5em minus 0.4em\relax IEEE, 2012, pp. 1--6.

\bibitem{hawes2017strands}
N.~Hawes, C.~Burbridge, F.~Jovan, L.~Kunze, B.~Lacerda, L.~Mudrova, J.~Young,
  J.~Wyatt, D.~Hebesberger, T.~Kortner, \emph{et~al.}, ``The strands project:
  Long-term autonomy in everyday environments,'' \emph{IEEE Robotics \&
  Automation Magazine}, vol.~24, no.~3, pp. 146--156, 2017.

\bibitem{hanheide2017and}
M.~Hanheide, D.~Hebesberger, and T.~Krajn{\'\i}k, ``The when, where, and how:
  An adaptive robotic info-terminal for care home residents,'' in \emph{HRI},
  2017, pp. 341--349.

\bibitem{alcantarilla2018street}
P.~F. Alcantarilla, S.~Stent, G.~Ros, R.~Arroyo, and R.~Gherardi, ``Street-view
  change detection with deconvolutional networks,'' \emph{Autonomous Robots},
  vol.~42, pp. 1301--1322, 2018.

\bibitem{palazzolo2018fast}
E.~Palazzolo and C.~Stachniss, ``Fast image-based geometric change detection
  given a 3d model,'' in \emph{ICRA}.\hskip 1em plus 0.5em minus 0.4em\relax
  IEEE, 2018, pp. 6308--6315.

\bibitem{wald2019rio}
J.~Wald, A.~Avetisyan, N.~Navab, F.~Tombari, and M.~Nie{\ss}ner, ``Rio: 3d
  object instance re-localization in changing indoor environments,'' in
  \emph{ICCV}, 2019, pp. 7658--7667.

\bibitem{lu20253dgs}
Z.~Lu, J.~Ye, and J.~Leonard, ``3dgs-cd: 3d gaussian splatting-based change
  detection for physical object rearrangement,'' \emph{IEEE Robotics and
  Automation Letters}, 2025.

\bibitem{sam}
A.~Kirillov, E.~Mintun, N.~Ravi, H.~Mao, C.~Rolland, L.~Gustafson, T.~Xiao,
  S.~Whitehead, A.~C. Berg, W.-Y. Lo, \emph{et~al.}, ``Segment anything,'' in
  \emph{ICCV}, 2023, pp. 4015--4026.

\bibitem{clip}
A.~Radford, J.~W. Kim, C.~Hallacy, A.~Ramesh, G.~Goh, S.~Agarwal, G.~Sastry,
  A.~Askell, P.~Mishkin, J.~Clark, \emph{et~al.}, ``Learning transferable
  visual models from natural language supervision,'' in \emph{ICML}.\hskip 1em
  plus 0.5em minus 0.4em\relax PmLR, 2021, pp. 8748--8763.

\bibitem{effsam}
Y.~Xiong, B.~Varadarajan, L.~Wu, X.~Xiang, F.~Xiao, C.~Zhu, X.~Dai, D.~Wang,
  F.~Sun, F.~Iandola, \emph{et~al.}, ``Efficientsam: Leveraged masked image
  pretraining for efficient segment anything,'' in \emph{CVPR}, 2024, pp.
  16\,111--16\,121.

\bibitem{ai2thor}
E.~Kolve, R.~Mottaghi, W.~Han, E.~VanderBilt, L.~Weihs, A.~Herrasti, D.~Gordon,
  Y.~Zhu, A.~Gupta, and A.~Farhadi, ``{AI2-THOR: An Interactive 3D Environment
  for Visual AI},'' \emph{arXiv preprint arXiv:1712.05474}, 2017.

\bibitem{moghaddam2021optimistic}
M.~K. Moghaddam, Q.~Wu, E.~Abbasnejad, and J.~Shi, ``Optimistic agent: Accurate
  graph-based value estimation for more successful visual navigation,'' in
  \emph{WACV}, 2021, pp. 3733--3742.

\bibitem{he2024relation}
Y.~He and K.~Zhou, ``Relation-wise transformer network and reinforcement
  learning for visual navigation,'' \emph{Neural Computing and Applications},
  vol.~36, no.~21, pp. 13\,205--13\,221, 2024.

\bibitem{icomma}
Y.~Sun, X.~Wang, Y.~Zhang, J.~Zhang, C.~Jiang, Y.~Guo, and F.~Wang, ``icomma:
  Inverting 3d gaussian splatting for camera pose estimation via comparing and
  matching,'' \emph{arXiv preprint arXiv:2312.09031}, 2023.

\end{thebibliography}

\end{document}